# INTERLANGUAGE SIGNS AND LEXICAL TRANSFER ERRORS

Atle Ro
Department of Linguistics and Phonetics
University of Bergen
ro@hf.uib.no

## Abstract
A theory of interlanguage (IL) lexicons is outlined, with emphasis on IL lexical entries, based on the HPSG notion of lexical sign. This theory accounts for idiosyncratic or lexical transfer of syntactic subcategorisation and idioms from the first language to the IL. It also accounts for developmental stages in IL lexical grammar, and grammatical variation in the use of the same lexical item. The theory offers a tool for robust parsing of lexical transfer errors and diagnosis of such errors.

## Introduction

Computational error diagnosis of inter language input should be based on second language acquisition theory. In this paper, I outline a theory of interlanguage lexical items which can be exploited computationally for robust parsing and error diagnosis. Interlanguage (IL) is interpreted in the sense of Selinker (1972). A unification-based lexicalist view of grammar is assumed (HPSG theory). The paper has the following outline: first lexical transfer is presented, then IL lexical entries are discussed, and finally some problems concerned with robust parsing and error diagnosis based on such lexical entries are pointed out.

## Lexical transfer

A theory of IL should account for lexical transfer from the first language (L1). By lexical transfer I mean that idiosyncratic properties of L1 lexical items are projected onto the corresponding target language (Lt) lexical items. By 'corresponding' I mean translationally related. I will consider two types of lexical transfer; translational transfer of idiomatic expressions, and transfer of L1 subcategorisation frames. Translational transfer of L1 idiomatic expressions is exemplified in (1), with translational transfer from French (Cf. Catt (1988)).

(1) *My friend has hunger.
    Mon ami a faim.

In unification-based grammatical frameworks, like HPSG, idiomatic expressions like *avoir faim* can be formalised as special cases of subcategorisation (for an LFG-style account of idioms, see Dyvik (1990))[1], and can thus be covered by the account I give of this type of transfer, to which we will now turn. The examples in (2)-(3) illustrate negative lexical transfer from Spanish in Norwegian interlanguage.

(2) *Jeg kunne ikke svare til Per.
    I could not answer to Per.

(3) No podía responder a Per.
    Not could-1sg answer to Per.
    'I could not answer Per'.

The Norwegian verb *svare* subcategorises for an object NP, while the Spanish verb with the same meaning, *responder*, subcategorises for a PP headed by the preposition *a* ("to"). The Spanish subcategorisation frame thus admits the erroneous VP (w.r.t. Norwegian grammar) in (2). I assume that IL lexical items are linked to corresponding L1 items by structure-sharing. Thus subcategorisation information from L1 lexical items can be used in generating interlanguage strings, and lexical transfer of the kind illustrated in (2) can be accounted for. I further make the idealisation that L1 word forms are not used as "terminal vocabulary" in interlanguage strings.

## Lexical entries in IL

How can these ideas be formalised? We will first consider the format which will be used for representing L1 lexical entries, that of

---

[1] Although it is not as straightforward as ordinary subcategorisation, one needs e.g. to distinguish between "pseudo-idioms" like *avoir faim* and real idioms like *kick the bucket*, but a discussion of this topic is not possible within this limited format.

Pollard and Sag (1987), and then outline interlanguage lexical entries. Example (4) illustrates an underspecified lexical entry for the transitive Norwegian verb *svare* ("answer").

(4)

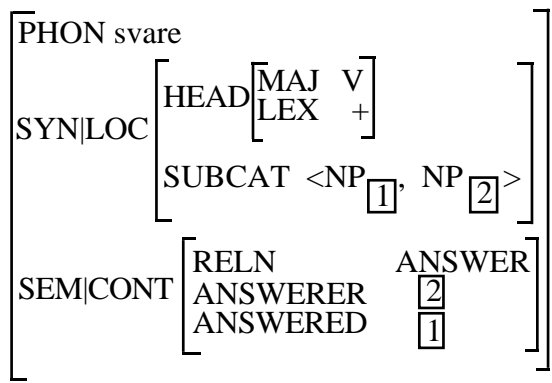

In HPSG, syntactic categories are signs and are represented as attribute-value matrices, with three features, PHON(OLOGY)[2], SYN(TAX) and SEM(ANTICS), which (with the exception of PHON) can have feature structures as values. The path SYN|LOC|SUBCAT takes a list of categories as value. The leftmost element in this list corresponds to the lexical head's most oblique complement (in (4) the object NP), the rightmost to the least oblique complement, (the subject in (4)). The feature SEM in this simple example is specified for the semantic relation expressed by the verb, as well as the roles the verb selects. The categories in the subcategorisation list are signs, for which the labels in (4) are a shorthand notation. The indices in the subcategorisation list indicate that the complement signs' SEM values are to bind the variables with which they are coindexed, and which are associated with the semantic roles in the relation. The lexical entry for the Spanish verb corresponding to *svare*, *responder*, is illustrated below:

(5)

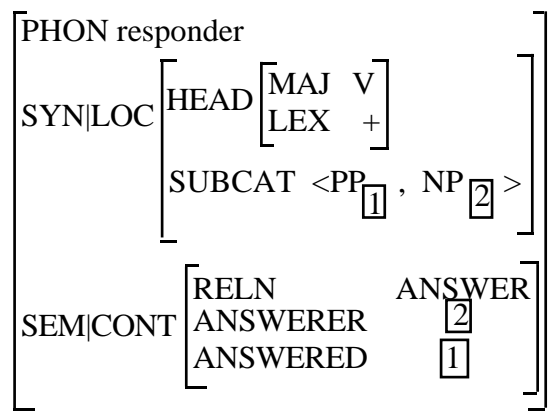

The signs in (4) and (5) are rather similar, with the exception of the first elements in the subcategorisation lists.[3]

(6)

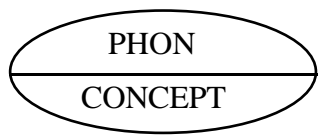

L1 sign

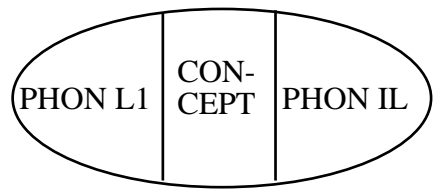

IL sign

The intuition is that whereas an L1 sign (in the sense of Saussurean sign[4]) has two sides, a concept side (corresponds to the feature SEM in HPSG) and an expression side (here called PHON), an IL sign has three sides: a common

---

[2] I follow Pollard and Sag (1987) and, for convenience, represent the value of PHON orthographically.

[3] Spanish direct objects are NPs when they refer to non-human entities, while objects which refer to humans must be expressed as PPs headed by the preposition *a* ("to"). It might appear to be a problem that NP and PP signs are of the same semantic type, but I follow Dyvik (1980) and Lødrup (1989) and call prepositional phrases which have their semantic role assigned by an external head, nominal; as opposed to modifier (adjunct) PPs, which express their own semantic roles. Having the same semantic type, "human" direct object PPs can be derived from NP objects by a lexical rule.

[4] Cf. de Saussure (1915).

concept side, a L1 expression side and an Lt (IL) expression side (cf. (6)).

Let us now return to the HPSG format, where it is possible to represent an IL entry and its corresponding L1 entry as a bilingual sign, similar to the concept used in machine translation (cf. Tsujii and Fujita (1991)). Interlanguage lexical entries can have the form which is visualised in (7)-(8), which represents two different alternatives and two different steps in the second language acquisition process:

(7)

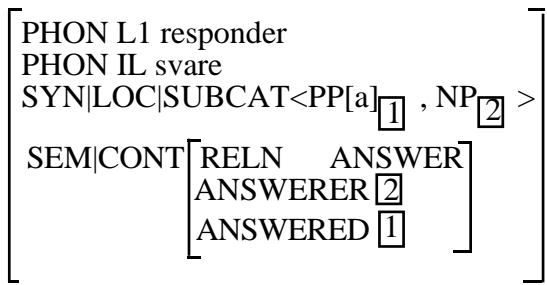

In this first alternative the expression side (PHON) of the IL entry is connected with its corresponding L1 entry. In this way lexical transfer from the L1 (cf. example (2)) is accounted for. The assumption is that the L1 lexical entry is basic at this stage, and the IL PHON is attached to it.

In a later developmental stage, where the syntactic properties of the IL entry are different from that of the corresponding L1 entry, the SYN|LOC|SUBCAT path of the IL entry (abbreviated below to SUBCAT) is given its own distinct value. This is illustrated in figure (8):

(8)

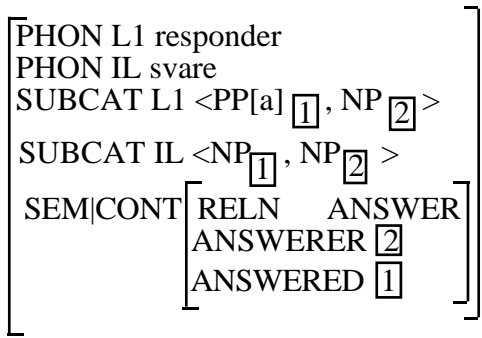

The lexical items are still linked because they have the same meaning.

The above account of lexical entries allows us to implement different stages in the development of an interlanguage lexicon. What it does not do, is to account for linguistic variability, which in L2 acquisition research generally is considered to be a property of L2 (cf. e.g. Larsen-Freeman and Long (1991)). In the case of lexical entries like (7) and (8), variability means that L2 users sometimes will use this item in accordance with Lt grammar, and sometimes not. But if we imagine a combination of (7) and (8), with a disjunctive IL subcat value, as illustrated in figure (9), variability is also catered for.

(9)

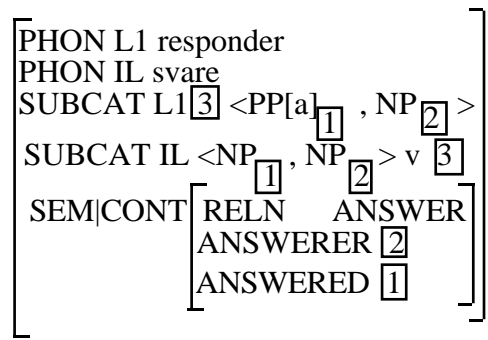

What we have just discussed also illustrates a significant difference between first and second language acquisition. As Odlin (1989) points out: " there is little reason to believe that the acquisition of lexical semantics in second language is a close recapitulation of all earlier development processes". Here we have an explication of this. Acquisition of lexical items is understood as associating Lt expressions with L1 concepts, either in a one-to-one relationship when L1 and Lt concepts are synonymous, or in one-to many relations when L1 and Lt concepts partially overlap[5].

## Future work

The prospect of the present approach to IL lexical items, lies in the possibility which a formalisation in a lexicalist, unification-based grammatical framework like HPSG gives for computational implementation, for purposes like anticipatory robust parsing and error diagnosis.

---

[5] In cases where an Lt lexical item not even partially corresponds to any L1 concept (e.g Japanese *bushido*, "the way of the samurai", if Japanese is Lt and English L1) the meaning of this item can still be paraphrased by means of L1 concepts.

The theory itself can be used for such purposes without much revision: the most important one is that in the bilingual signs IL attribute-value pairs must be replaced with Lt ones. A system for robust parsing and error diagnosis of lexical transfer errors needs Lt subcat values for determining whether sentences have the right complements or not, and L1 subcat values for making hypotheses about possible erroneous complements.

My idea is to exploit the relation between Lt and L1 subcat values in a chart parser, so if a parse fails because of a mismatch between a complement in the input string and the Lt complement needed by a lexical head, the chart can be modified such that a complement licensed by the L1 subcat specification is introduced into the chart as an alternative to the incompatible complement specification in the Lt subcat list.

This is not unproblematic, however, because even in successful chart parses, many hypotheses fail. Lexical entries with alternative subcategorisation frames (e.g. SUBCAT <X,(Y),(Z)>, where Y and Z are optional) are common. The longer the sentence, and the more lexical heads, the larger is the number of hypotheses which will fail. In the case of erroneous strings, how can a system decide which hypotheses to modify, in order to accept such strings? Modifying all edges which fail will even in simple sentences soon lead to an explosion of new edges. Deciding which error hypotheses are the most promising is a central computational question to which future work must be dedicated.

## Conclusion

A theory of IL signs, which accounts for lexical transfer, has been presented. Transfer has been a controversial subject in second language acquisition (SLA) research, and its importance as a property of L2s has been evaluated differently through the history of SLA research. Scientists have disagreed whether properties of L2 production can be attributed to transfer or other factors, such as universal developmental sequences (cf. e.g. Meisel et al. (1981)). In my opinion lexical transfer is an interesting aspect of this discussion. As it is associated with individual lexical items it has a stronger case that more general types of transfer (e.g. transfer of word order). When faced with an error where syntactic or semantic properties of an IL lexical item diverge from the standard of Lt, but are strikingly similar to properties of the corresponding L1 lexical item, lexical transfer is a likely explanation.

## Acknowledgements

I would like to thank Helge Dyvik, Torbjørn Nordgård and two anonymous reviewers for their valuable comments. I also want to thank Richard Pierce for his helpful advice with the wording of the paper.

## References

Catt, M. E. (1988). *Intelligent Diagnosis of Ungrammaticality in Computer-Assisted Language Instruction.* Technical Report CSRI-218. University of Toronto.

Dyvik, H. (1980). *Grammatikk og empiri.* Doctoral thesis, University of Bergen.

Dyvik, H. (1991). *The PONS Project. Features of a Translational System.* Skrifserie nr. 39, Department of Linguistics and Phonetics, University of Bergen.

Larsen-Freeman, D. and M. H. Long (1991). *An Introduction to Second Language Acquisition Research.* Longman, London.

Lødrup, H. (1989). *Norske hypotagmer. En LFG-beskrivelse av ikke-verbale norske hypotagmer.* Novus, Oslo.

Meisel, J, H. Clahsen and M. Pienemann (1981). On determining developmental stages in natural language acquisition. *Studies in second language acquisition* 3 (1): 109-35.

Odlin, T. (1989). *Language Transfer. Cross-linguistic influence in language learning.* Cambridge University Press, Cambridge.

Pollard, C. and I. T. Sag (1987). *Information-Based Syntax and Semantics. Volume 1: Fundamentals.* CSLI Lecture Notes, Stanford.

Saussure, F. de (1915). *Course in General Linguistics.* Published in 1959 by MacGraw Hill, New York.

Selinker, L. (1972). Interlanguage. In *IRAL*, X.3, 219-231.

Tsujii, J. and K. Fujita (1991). Lexical transfer based on bilingual signs: Towards interaction during transfer. In *Proceedings of EACL*, Berlin, Germany.